\documentclass[runningheads]{llncs}
\usepackage{graphicx}
\usepackage{mathptmx}
\usepackage{graphicx}
\usepackage{times}
\usepackage{csquotes}
\usepackage{comment}
\usepackage{float}
\usepackage{enumitem}
\usepackage{stfloats}
\usepackage{booktabs}

\usepackage{listings}
\usepackage{caption}
\usepackage{subcaption}
\usepackage{multicol}
\usepackage[table,xcdraw]{xcolor}
\usepackage[utf8]{inputenc}
\usepackage{longtable}
\usepackage{multirow}
\usepackage{hhline}
\usepackage{hyperref}
\newcolumntype{L}{>{\centering\arraybackslash}m{3cm}}

\DeclareGraphicsExtensions{.jpg,.png}

\begin{document}
\title{HR-Crime: Human-Related Anomaly Detection in Surveillance Videos}
%Information extraction from crime surveillance videos for human-related anomaly detection \thanks{Supported by organization x.}


\author{Kayleigh Boekhoudt\inst{1}\inst{\dag} \and
Alina Matei\inst{1} \and
Maya Aghaei\inst{2}\inst{*} \and
Estefanía Talavera\inst{1}\inst{3}\inst{*}}
\authorrunning{K. Boekhoudt et al.}

\institute{University of Groningen, Groningen, the Netherlands \and 
NHL Stenden University of Applied Sciences, Leeuwarden, the Netherlands \and University of Twente, Enschede, the Netherlands \\
\inst{* } Authors contributed equally
\\
\inst{\dag} Corresponding author \{k.j.boekhoudt@student.rug.nl\}}

\maketitle              % typeset the header of the contribution
\begin{abstract}
%The abstract should briefly summarize the contents of the paper in
The automatic detection of anomalies captured by surveillance settings is essential for speeding the otherwise laborious approach. To date, UCF-Crime is the largest available dataset for automatic visual analysis of anomalies and consists of real-world crime scenes of various categories. In this paper, we introduce \textit{HR-Crime}, a subset of the UCF-Crime dataset suitable for \textit{human-related} anomaly detection tasks. We rely on state-of-the-art techniques to build the feature extraction pipeline for human-related anomaly detection. Furthermore, we present the baseline anomaly detection analysis on the HR-Crime. HR-Crime as well as the developed feature extraction pipeline and the extracted features will be publicly available for further research in the field.

\end{abstract}

%\textcolor{red}{link to paper before moving to main everything :} %https://www.overleaf.com/5672348439smrcnwfzqktd}

\section{Introduction}
\label{Section:Introduction}

%what is out there:
The detection of anomalous events in videos is a challenging task due to the broad definition of the term `anomaly', as well as insufficient annotated data. Despite this, there has been much research in the field of video surveillance anomaly detection in the past years \cite{video_anomaly_detection_survey}. Surveillance cameras are a widely used technology which aids law enforcement agencies in ensuring general public safety. Surveillance footage is also considered a reliable piece of forensic evidence when the anomalies captured on the footage are identified as crimes. However, due to the overwhelming amount of surveillance video data (usually, surveillance cameras transmit 24/7 live footage), there is an outstanding need for the automation of abnormality detection in such videos.

\begin{figure}[hbt!]
    \centering
    \includegraphics[width = .9\textwidth, trim={0 4.5cm 0 3.6cm}, clip]{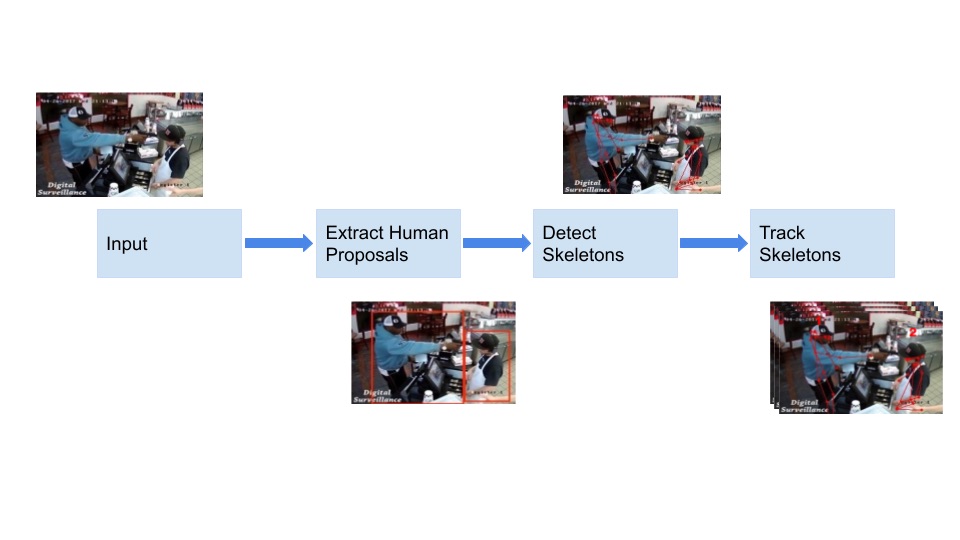}
    \caption{Feature extraction pipeline of HR-Crime dataset. Given the frames of a video, we first extract human proposals using YOLOv3-spp \cite{YOLOv3-spp}. Second, AlphaPose \cite{alphapose} is applied to detect body skeletons. Finally, PoseFlow \cite{poseflow} is used to track skeletons.}
    \label{fig:pipeline}
\end{figure}

Human-related anomaly detection in surveillance videos, as a specific case of anomaly detection, is closely related to human activity detection that can be visually recognized as abnormal through body movement. In recent years, there have been many advances in the field of human pose (skeleton) estimation. Generally, there are two main types of frameworks used for pose detection. Two-step frameworks such as AlphaPose \cite{alphapose} use the top-down approach for pose detection. The idea is to first use an object detector to find people and then perform single-person pose estimation for each detected person. In contrast, methods that use the bottom-up approach to detect poses, first localize body parts and then group them into person instances \cite{openpose} \cite{metu}.

There are advantages and disadvantages to these methods. Bottom-up methods feed the whole image to their architecture, which may impose limitations on the input image size. On the other hand, top-down methods crop and feed each individually detected human bounding boxes to their architecture. The disadvantages of the top-down method is that the keypoint detection performance depends on the quality of the bounding boxes and that the runtime is proportional to the number of people in the frame. Bottom-up approaches, on the contrary, do not have the issue of early commitment and runtime complexity. Cao et al. \cite{openpose} suggests that AlphaPose \cite{alphapose} to be used for maximum accuracy, OpenPose \cite{openpose} for maximum speed, and METU \cite{metu} for a trade-off between them.

Many efforts have also been made in recent years towards accurate human pose tracking. PoseTrack \cite{posetrack} and ArtTrack \cite{arttrack} introduced the multi-person pose tracking incorporating the idea of the part-based pose estimator DeeperCut \cite{deepercut} by extending spatial joint graph to spatio-temporal graph. First, the model generates for each frame, a set of detected keypoints and constructs the spatio-temporal graph. The model then solves an integer linear program to divide the graph into sub-graphs that correspond to skeleton trajectories of each person. These methods are also known as jointing schemes and are computationally heavy and not scalable to long videos. Top-down methods such as PoseFlow \cite{poseflow} are more scalable. The model starts by detecting human bounding boxes in every frame and extracts poses from each bounding box. The boxes are then tracked over the entire video in terms of similarity between pairs of boxes. These types of pose trackers are therefore also called Detect-And-Track methods. PoseFlow \cite{poseflow} also takes motion and pose information into account by implementing a cross frame matching technique to propagate box proposals to previous and next frames. 

Extending on the task of video anomaly detection, small progress has been made targeting the human-related anomaly detection plainly \cite{human_anomaly} \cite{human_anomaly_1}. In \cite{skeleton_based_anomaly_detection}, the authors proposed the MPED-RNN architecture for anomaly detection in surveillance videos based on skeleton trajectories described by local and global body movement. The proposed MPED-RNN follows an encoder-decoder architecture: the encoder learns close approximations of normal trajectory which are decoded with high accuracy; this implies that, when presented with abnormal trajectories, the encoder-decoder architecture obtains inaccurate reconstructions which results in high anomaly scores.

%%%%%%%%%%%
% Contributions
%%%%%%%%%%%
One reason for the small progress in human-related anomaly analysis might be the lack of human centered anomaly related datasets. Hence, the main contributions of this work are planned to target this shortage as following: 
\begin{enumerate}
\item We introduce and make publicly available the Human-Related Crime dataset (HR-Crime) together with the annotations at the frame level. \footnote{Dataset is publicly available at \href{https://doi.org/10.34894/IRRDJE}{https://doi.org/10.34894/IRRDJE}}
\item We present baseline results on HR-Crime intending to contribute to future research in the field of human-related anomaly detection.
\end{enumerate} 

% PAPER ORGANIZATION
The rest of the paper is organized as follows. In Section \ref{Section:methodology}, we discuss our feature extraction pipeline for development of the HR-Crime dataset. In Section \ref{Section:results}, we elaborate on the implementation details and discuss the obtained results. Finally, we draw conclusions in Section \ref{Section:discussion}.

%The rest of the paper is organized as follows. In Section \ref{Section:methodology}, we discuss our feature extraction pipeline for development of the HR-Crime dataset. In Section \ref{Section:results}, we elaborate on the implementation details and discuss the obtained results. Later, we introduce the experiment designs in Section \ref{Section:results} and draw conclusions in Section \ref{Section:discussion}.

%%%%%%%% --------------------------------------
%%%%%%%% ------------ DATASET  --------
%%%%%%%% --------------------------------------
%%%%%%%% --------------------------------------

\section{HR-Crime dataset}\label{Section:methodology}

As mentioned earlier, Morais et al. recently introduced the only work on human-related anomaly detection in surveillance videos  \cite{skeleton_based_anomaly_detection}. Their introduced architecture, MPED-RNN, requires a defined set of features extracted from videos to detect the human-related abnormalities. In an attempt to provide the baseline results on HR-Crime, we opt for extracting the required features from the UCF-Crime \cite{UCF_Crime} videos and only keep the relevant information out of it to build the HR-Crime dataset. In this section, we describe the followed steps to prepare the HR-Crime dataset.

\subsection{HR-Crime statistics}
%data + quantification
The UCF-Crime dataset \cite{UCF_Crime} consists of 950 real-world surveillance videos of anomalies, and 950 normal videos. The anomalies are divided into 13 categories: \textit{Abuse}, \textit{Arrest}, \textit{Arson}, \textit{Assault}, \textit{Burglary}, \textit{Explosion}, \textit{Fighting}, \textit{Road Accidents}, \textit{Robbery}, \textit{Shooting}, \textit{Shoplifting}, \textit{Stealing} and \textit{Vandalism}. Duplicates may occur because some videos either have multiple anomalies or the anomaly may fall into more than one category. UCF-Crime dataset as is originally gathered to represent the \textit{in-the-wild} nature of the crime scenes, at times lacks the required clarity in content even for human eyes. This comes in addition to the fact that only a subset of it is human-related. Hence, for further human-related anomaly analysis, we extracted HR-Crime out of UCF-Crime dataset using the following guidelines:
\begin{itemize}[noitemsep,topsep=0pt]
    \item Omitting videos of anomalous events that are not human-related. We refer to `human-related' if the main performing subjects are human. Within this definition, dog abuse is not considered human-related.
    \item Excluding videos that do not have a clear view of the people at the scene.
    \item Ignoring videos with large crowds, as our goal is not to do crowd analysis which is essentially a different task than human behavior analysis.
    \item Ignoring videos longer than 100 minutes.
\end{itemize}

\begin{figure}[H]
     \centering
     \begin{subfigure}[b]{0.3\textwidth}
         \centering
         \includegraphics[width=\textwidth, height=2.5cm]{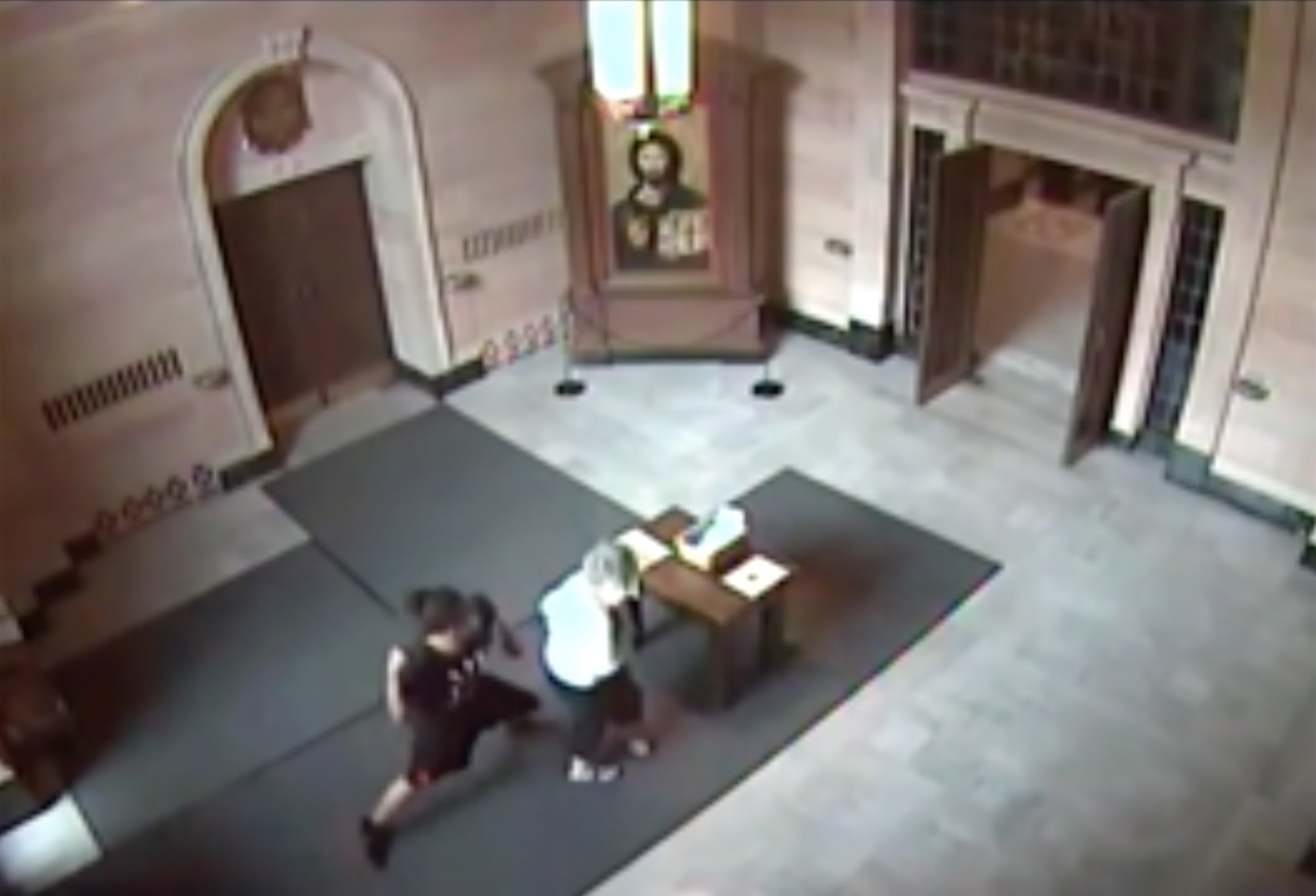}
         \caption{Abuse}
     \end{subfigure}
     \hspace{4 mm}
     \begin{subfigure}[b]{0.3\textwidth}
         \centering
         \includegraphics[width=\textwidth, height=2.5cm]{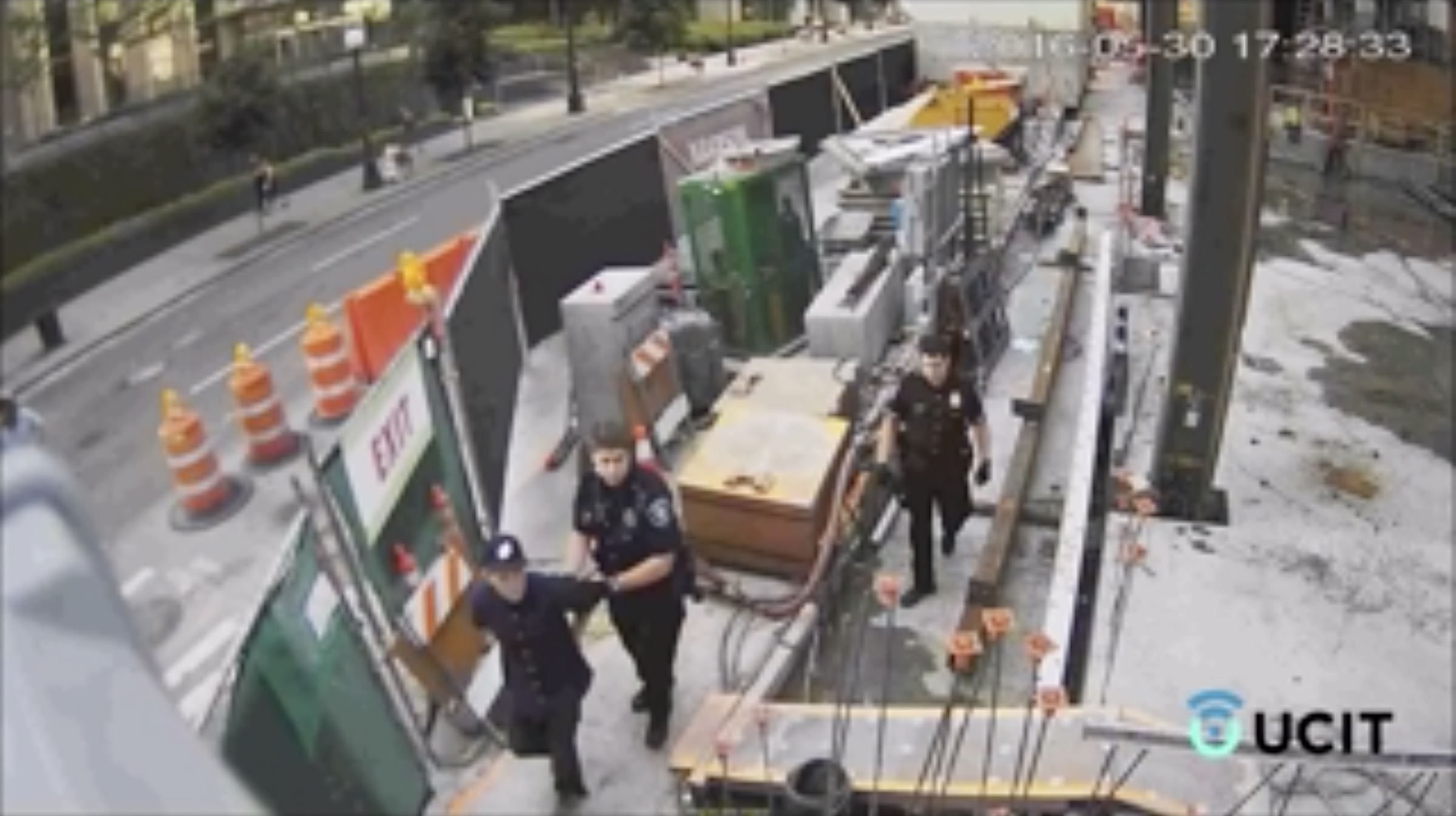}
         \caption{Arrest}
     \end{subfigure}
     \hspace{4 mm}
     \begin{subfigure}[b]{0.3\textwidth}
         \centering
         \includegraphics[width=\textwidth, height=2.5cm]{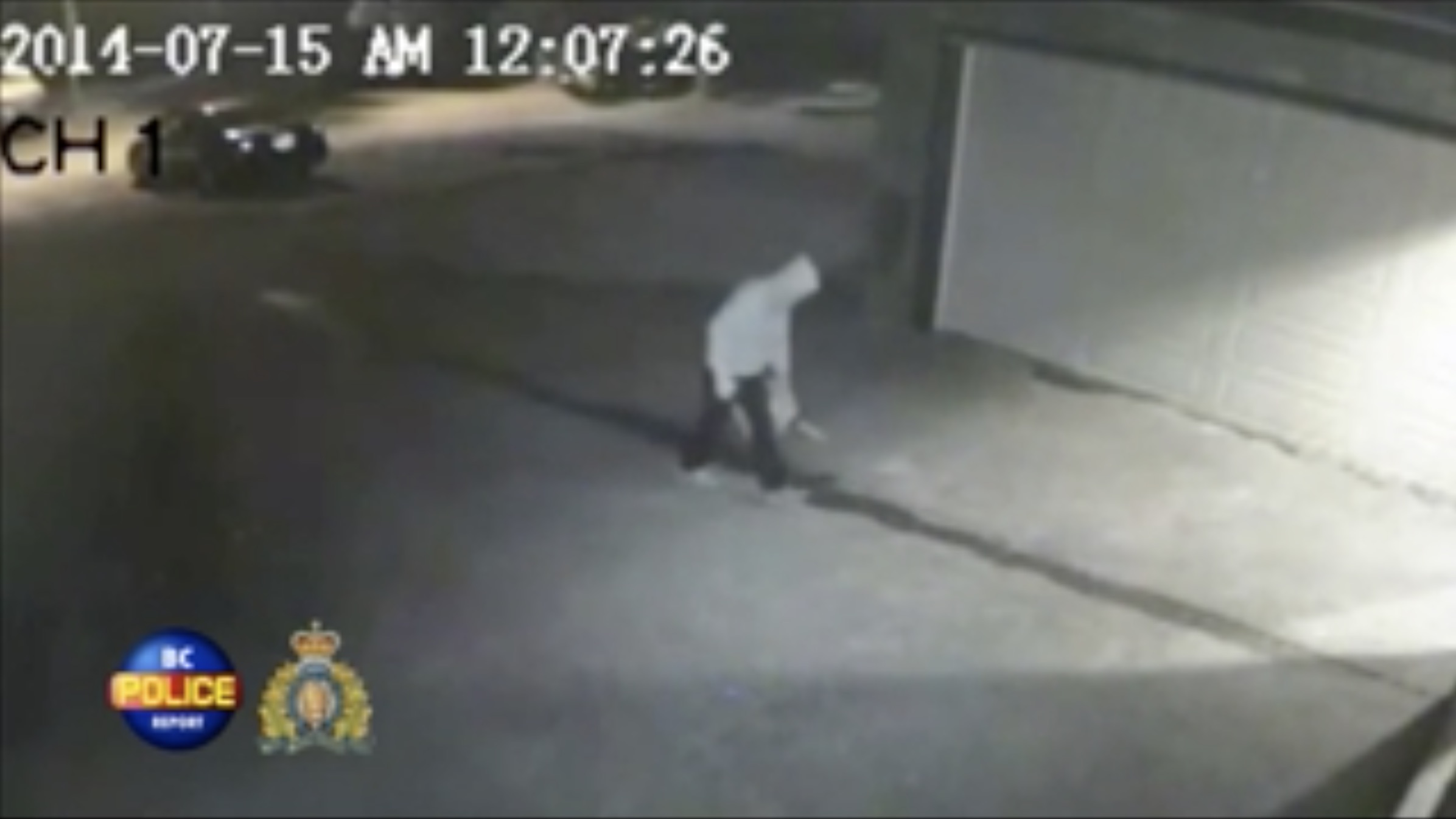}
         \caption{Arson}
     \end{subfigure}
     \newline
     %\vspace{0.5cm}
     \begin{subfigure}[b]{0.3\textwidth}
         \centering
         \includegraphics[width=\textwidth, height=2.5cm]{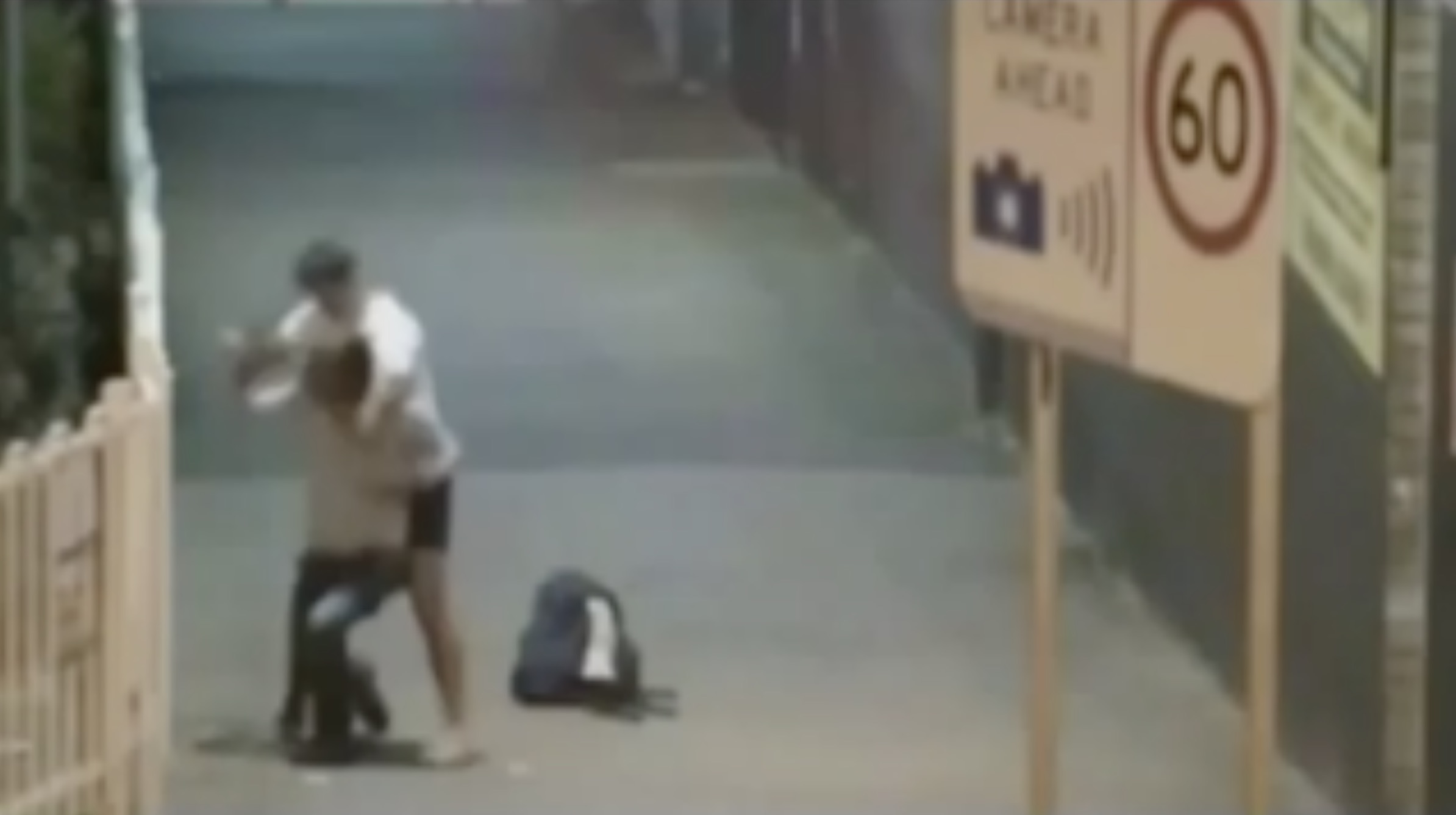}
         \caption{Assault}
     \end{subfigure}
     \hspace{4 mm}
     \begin{subfigure}[b]{0.3\textwidth}
         \centering
         \includegraphics[width=\textwidth, height=2.5cm]{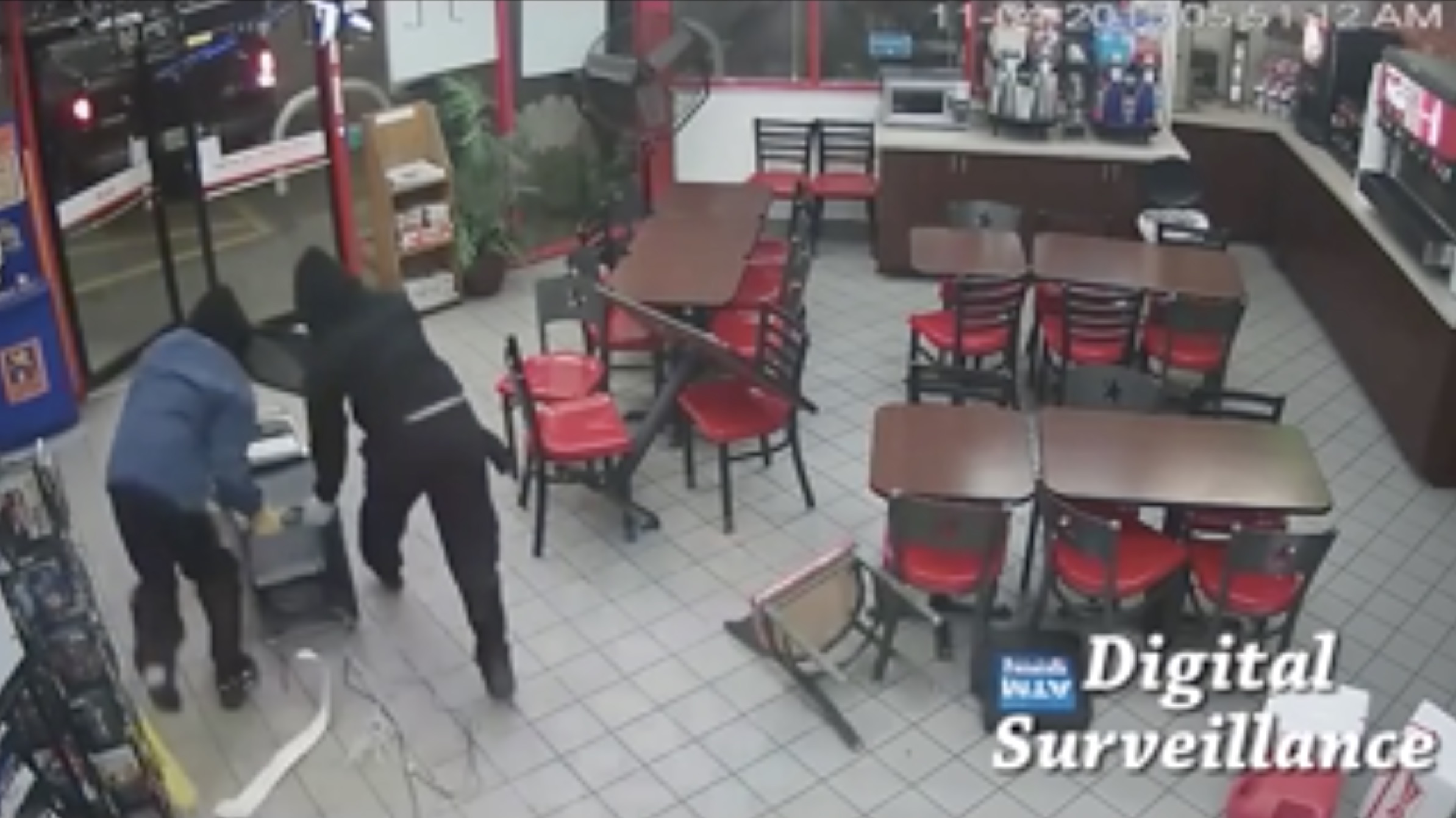}
         \caption{Burglary}
     \end{subfigure}
     \hspace{4 mm}
     \begin{subfigure}[b]{0.3\textwidth}
         \centering
         \includegraphics[width=\textwidth, height=2.5cm]{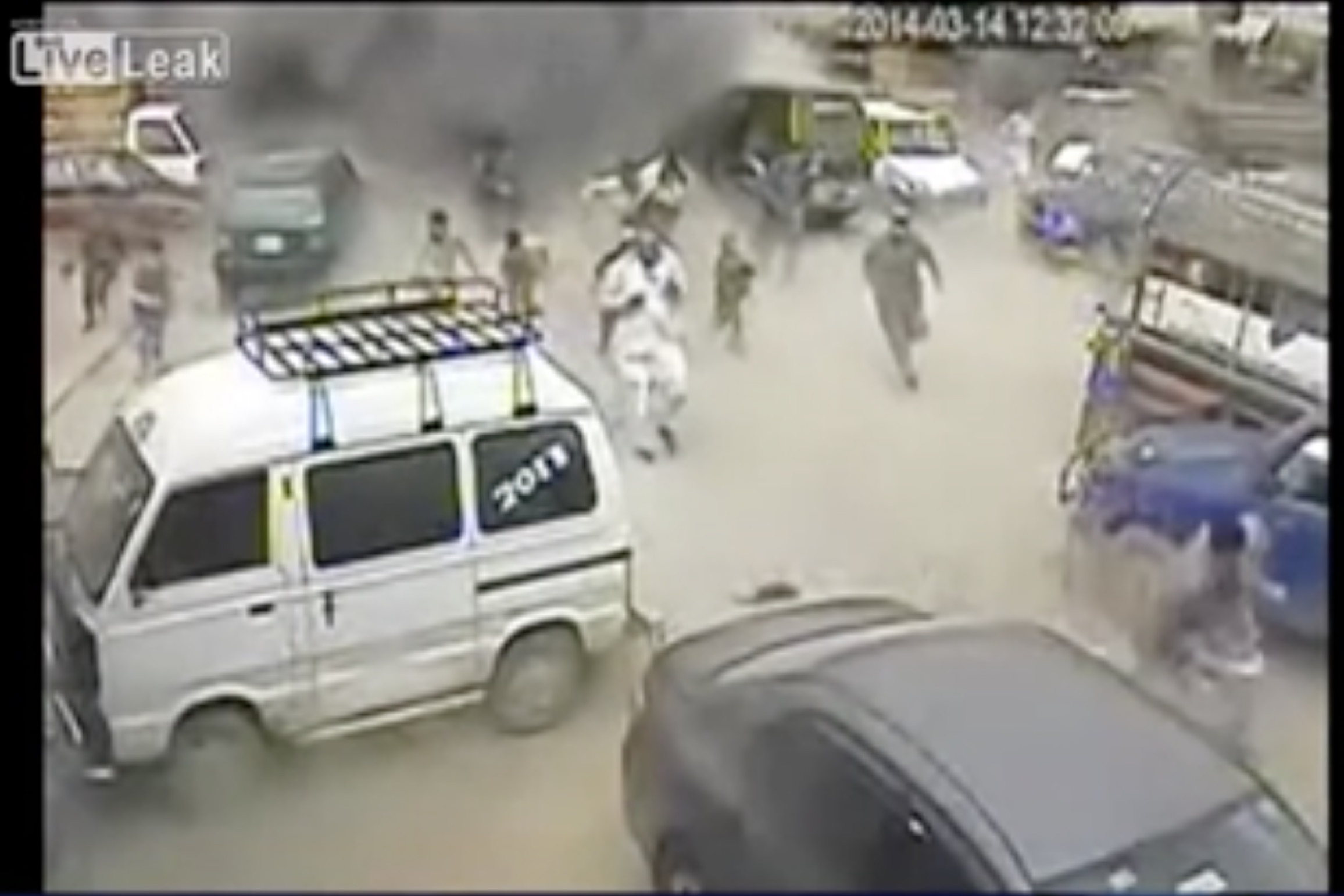}
         \caption{Explosion}
     \end{subfigure}
     \newline
     %\vspace{0.5cm}
     \begin{subfigure}[b]{0.3\textwidth}
         \centering
         \includegraphics[width=\textwidth, height=2.5cm]{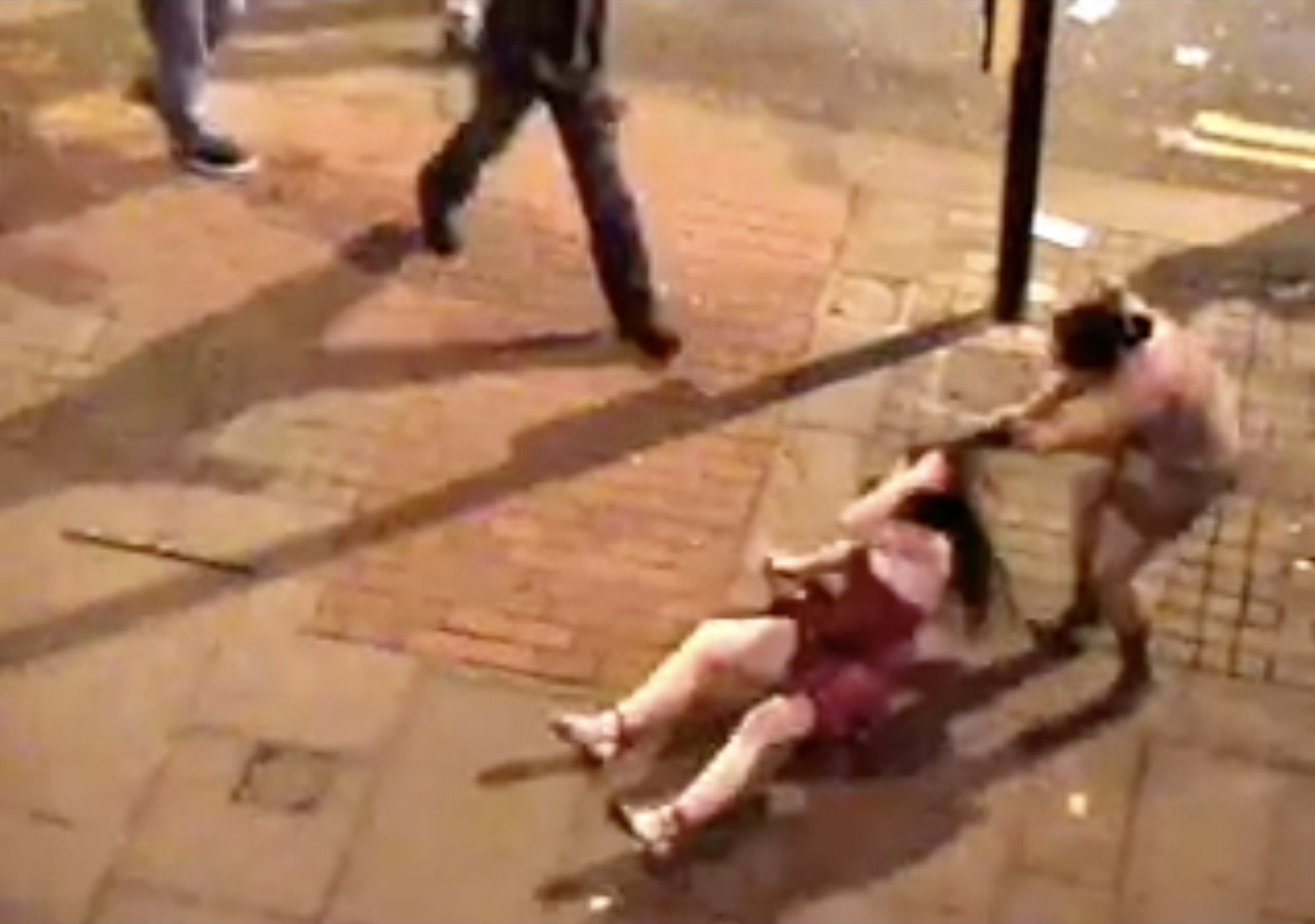}
         \caption{Fighting}
     \end{subfigure}
     \hspace{4 mm}
     \begin{subfigure}[b]{0.3\textwidth}
         \centering
         \includegraphics[width=\textwidth, height=2.5cm]{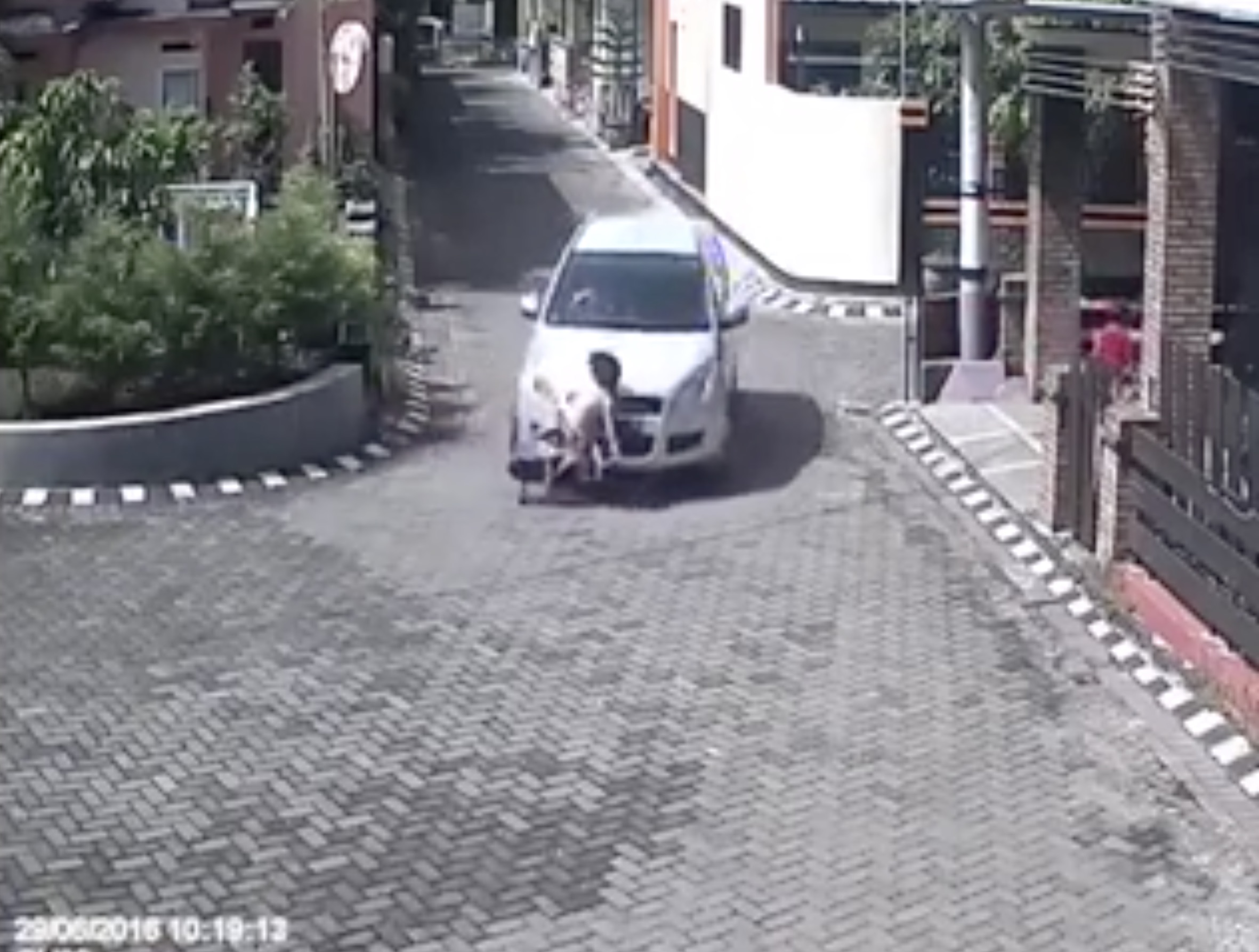}
         \caption{Road Accidents}
     \end{subfigure}
     \hspace{4 mm}
     \begin{subfigure}[b]{0.3\textwidth}
         \centering
         \includegraphics[width=\textwidth, height=2.5cm]{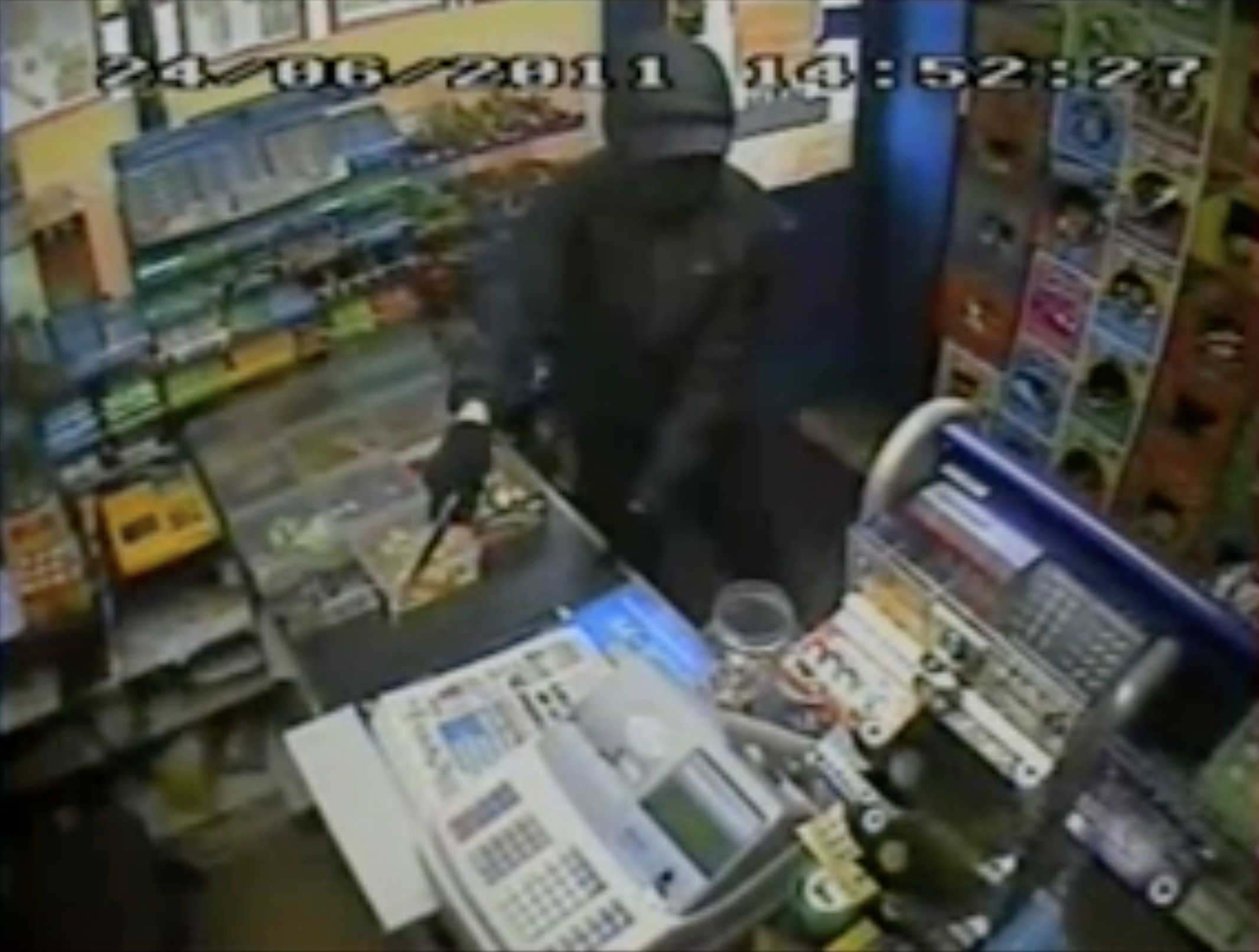}
         \caption{Robbery}
     \end{subfigure}
     \newline
     %\vspace{0.5cm}
     \begin{subfigure}[b]{0.3\textwidth}
         \centering
         \includegraphics[width=\textwidth, height=2.5cm]{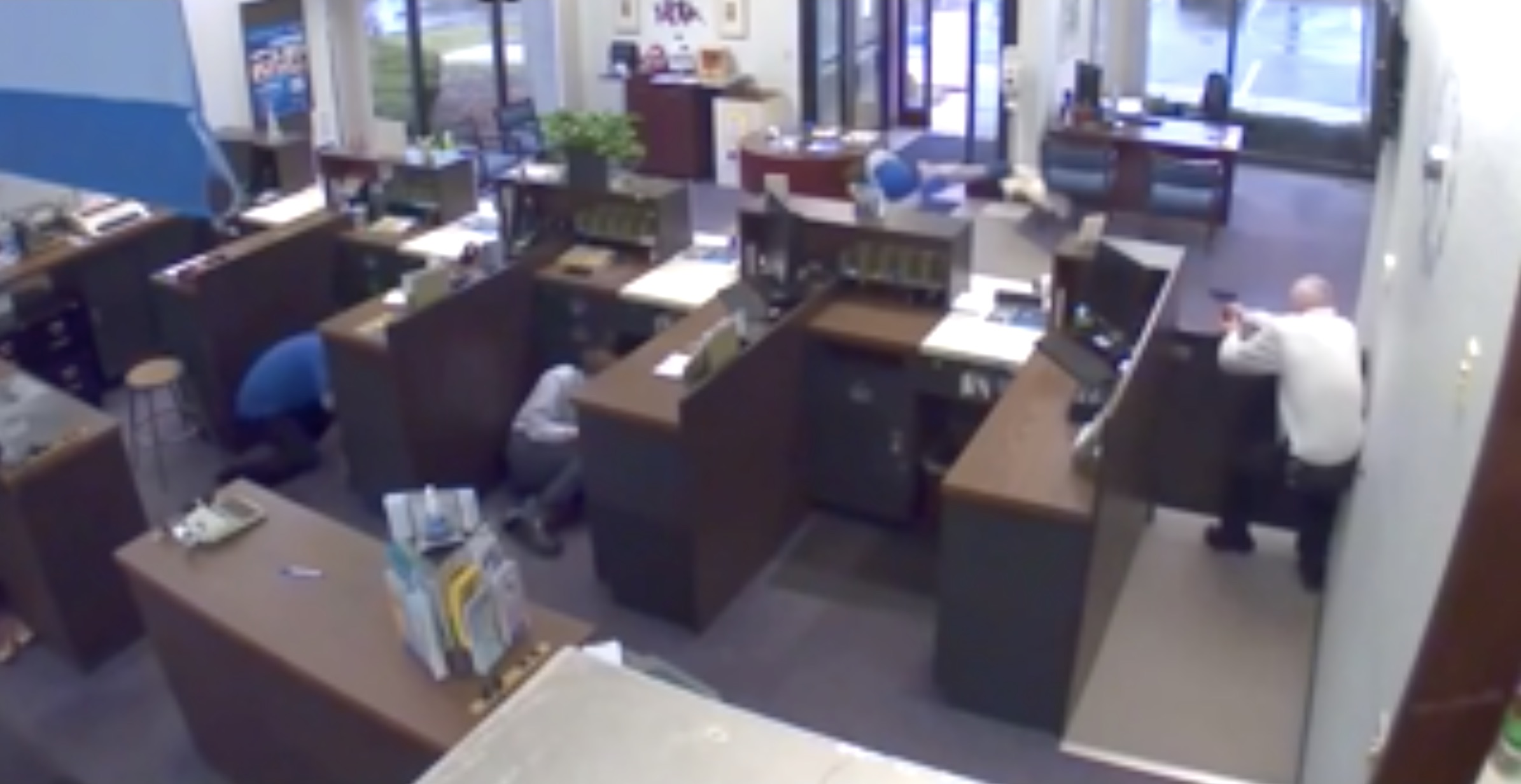}
         \caption{Shooting}
     \end{subfigure}
     \hspace{4 mm}
     \begin{subfigure}[b]{0.3\textwidth}
         \centering
         \includegraphics[width=\textwidth, height=2.5cm]{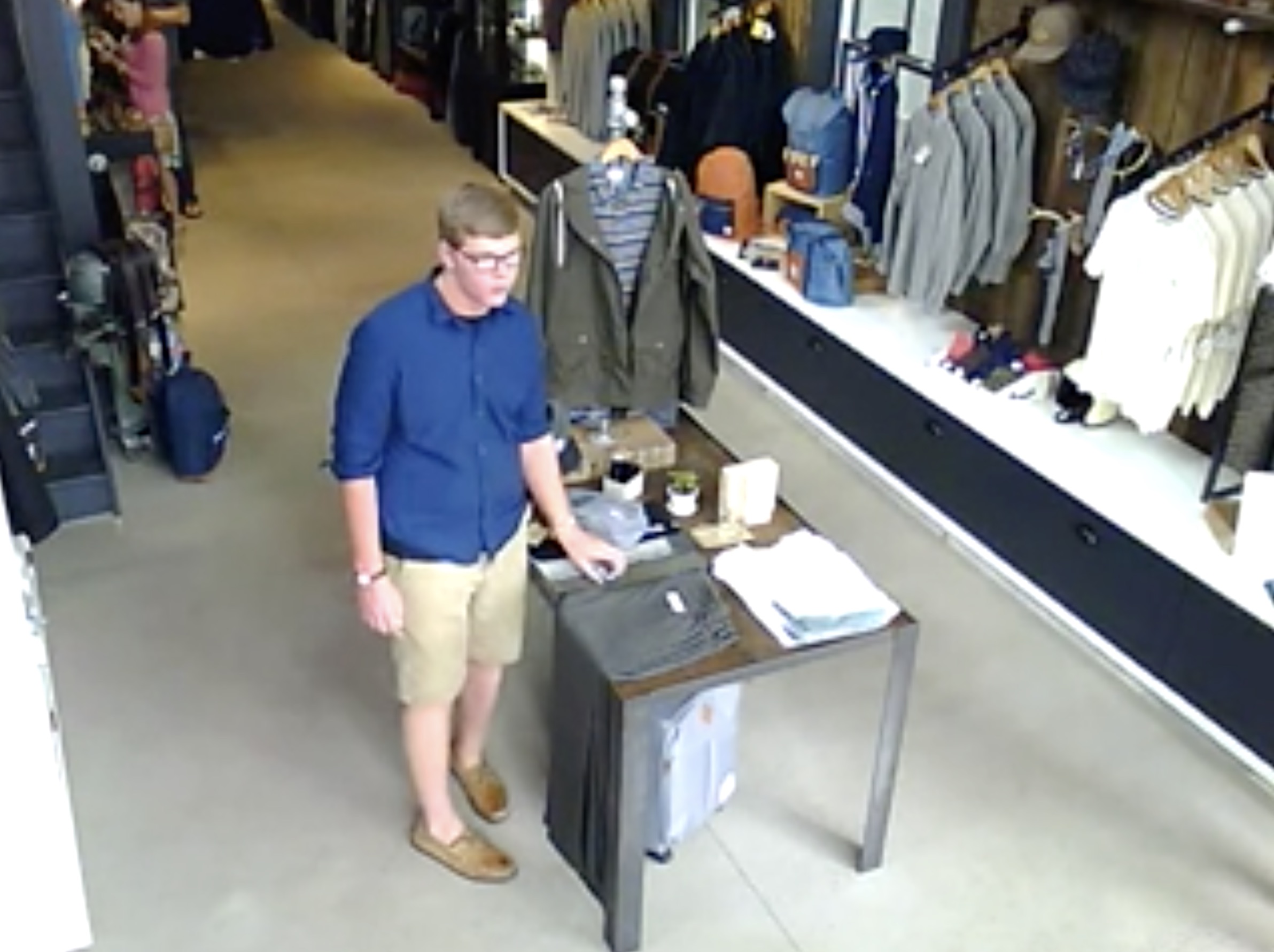}
         \caption{Shoplifting}
     \end{subfigure}
     \hspace{4 mm}
     \begin{subfigure}[b]{0.3\textwidth}
         \centering
         \includegraphics[width=\textwidth, height=2.5cm]{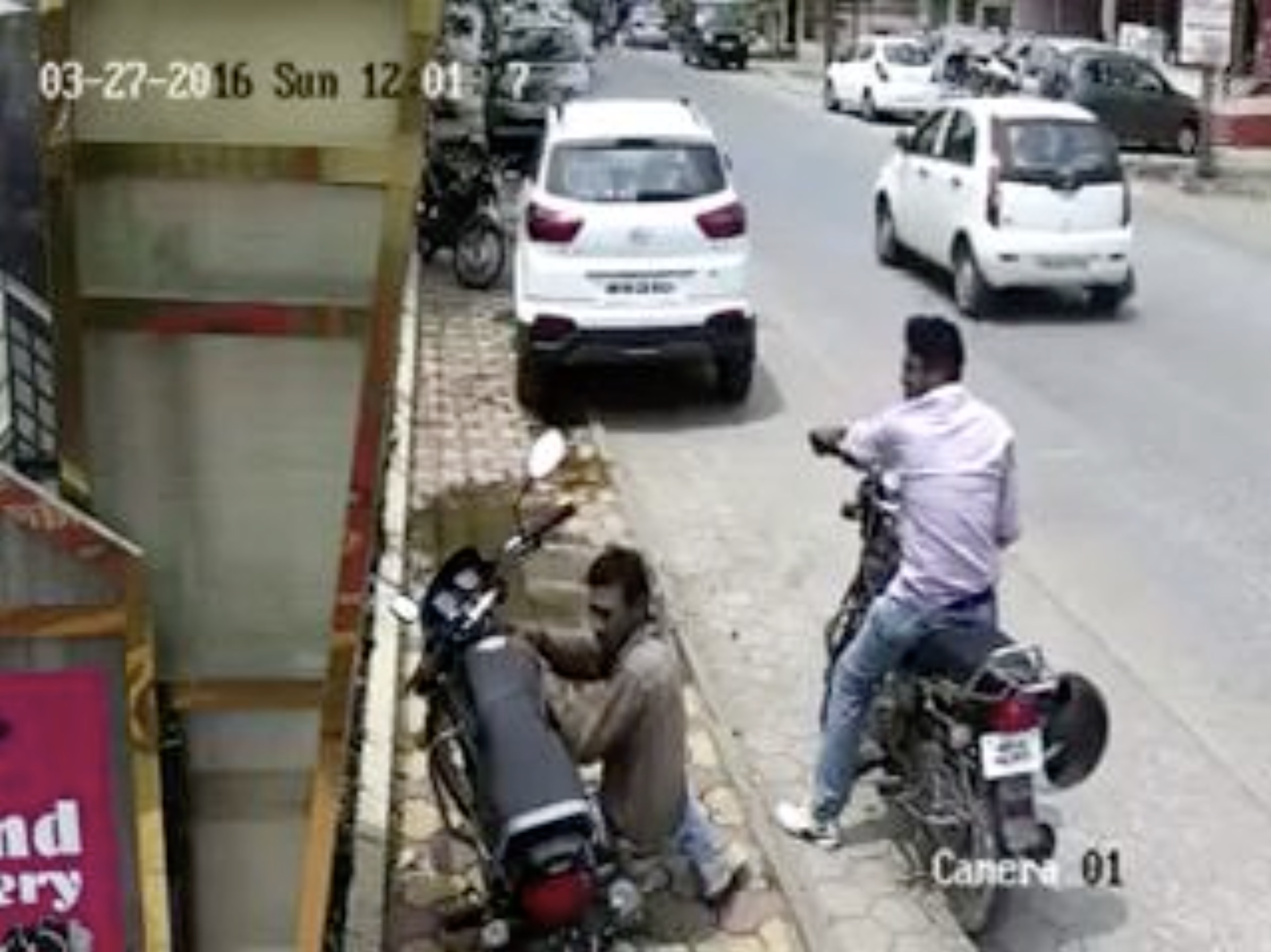}
         \caption{Stealing}
     \end{subfigure}
     \newline
     %\vspace{0.5cm}
     \begin{subfigure}[b]{0.3\textwidth}
         \centering
         \includegraphics[width=\textwidth, height=2.5cm]{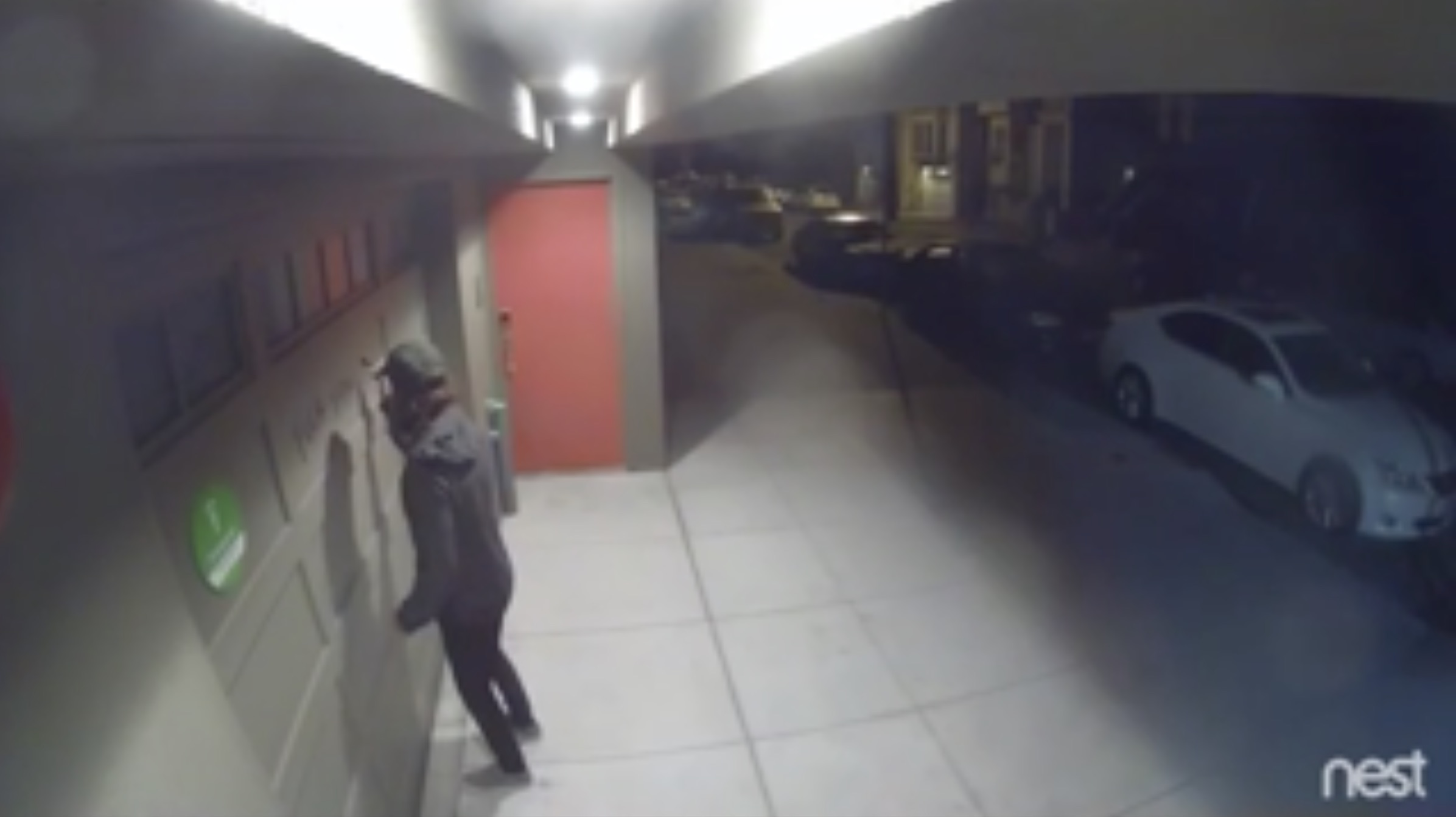}
         \caption{Vandalism}
     \end{subfigure}
     \hspace{4 mm}
     \begin{subfigure}[b]{0.3\textwidth}
         \centering
         \includegraphics[width=\textwidth, height=2.5cm]{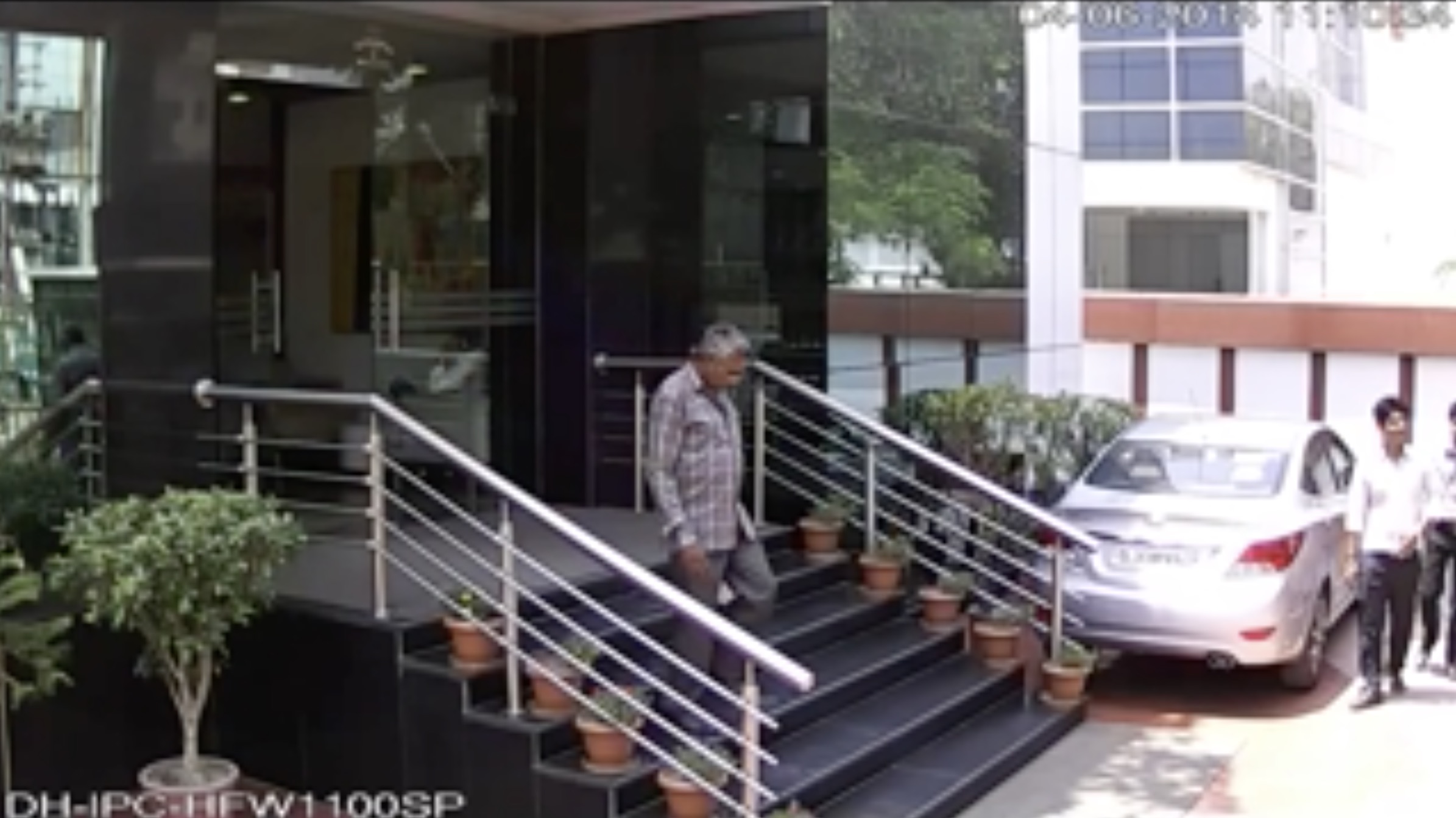}
         \caption{Normal}
     \end{subfigure}
     %\hspace{37mm}
     \hspace{0.3\textwidth}\hspace{4 mm}
     \caption{From (a) to (n), example frames from HR-Crime: a man punches an older woman; the police escort a man; a person pours inflammable liquid on the ground; a man holds another man by his neck; two burglars steal an ATM; people run away from an explosion; a woman drags another woman by her hair; a car hits a child on a bicycle; a robber holds a knife; a shooting in an office; a man shoplifts a watch; a man hot-wires a motorcycle; a vandal writes on a garage door; a man walks down some stairs.}
     \label{fig:crime_dataset_examples}
\end{figure}

\begin{figure}[H]
    \centering
    \begin{subfigure}[b]{\textwidth}
        \centering
        %trim={<left> <lower> <right> <upper>}
       \includegraphics[width=0.9\textwidth, trim={0 0 0 0}, clip]{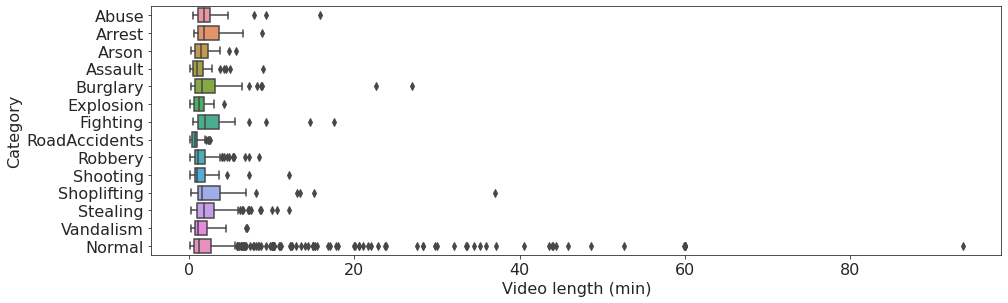}
        \caption{Video length in minutes}
        \label{fig:video_length}
    \end{subfigure}
    \newline
    \begin{subfigure}[b]{\textwidth}
        \centering
        \includegraphics[width=.9\textwidth, trim={0 0 0 0}, clip]{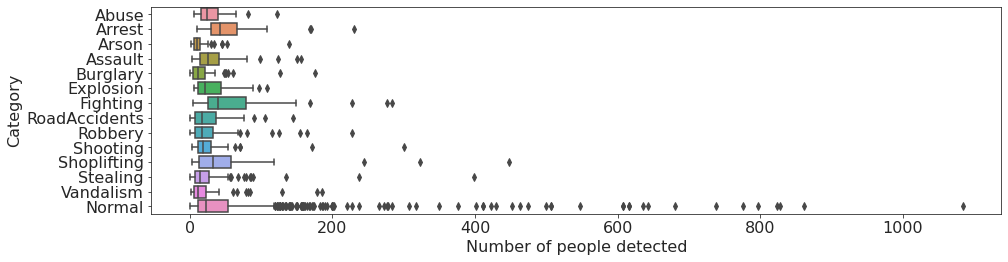}
        \caption{Skeleton trajectories}
        \label{fig:number_of_people_detected}
    \end{subfigure}
    \caption{Video length and number of skeleton trajectories distribution in HR-Crime.}
    \label{fig:my_label}
\end{figure}

\begin{table}[b!]
\centering
\resizebox{\textwidth}{!}{\begin{tabular}{p{2.5cm}||r|r|r|r|r|r|r|r|r|r|r|r|r|r|r}

 & \rotatebox{90}{Abuse} & \rotatebox{90}{Arrest} & \rotatebox{90}{Arson} & \rotatebox{90}{Assault} & \rotatebox{90}{Burglary} & \rotatebox{90}{Explosion} & \rotatebox{90}{Fighting} & \rotatebox{90}{\begin{tabular}[l]{@{}l@{}}Road\\ Accidents\end{tabular} } & \rotatebox{90}{Robbery} & \rotatebox{90}{Shooting} & \rotatebox{90}{Shoplifting} & \rotatebox{90}{Stealing} & \rotatebox{90}{Vandalism} & \rotatebox{90}{\begin{tabular}[l]{@{}l@{}}\textbf{Total}\\ \textbf{anomalous}\end{tabular} } & \rotatebox{90}{\begin{tabular}[l]{@{}l@{}}\textbf{Total}\\ \textbf{normal}\end{tabular}}\\
\hline
\hline
Videos & 50 & 50 & 50 & 50 & 100 & 50 & 50 & 150 & 150 & 50 & 50 & 100 & 50 & \textbf{950} & \textbf{950}\\
\hline
HR-videos & 38 & 42 & 48 & 47 & 96 & 26 & 39 & 68 & 145 & 46 & 50 & 98 & 46 & \textbf{789} & \textbf{782}\\
\hline
Test HR-videos & 0 & 5 & 9 & 3 & 13 & 12 & 5 & 13 & 5 & 21 & 21 &	5 &	5 & \textbf{117} & \textbf{122}\\
\hline
Anomalous frames & 0     &   7820  &   8166  &   8520  &   16104 &   3097  &   4437  &   1233 &   3815  &   9314  &   7525  &   6007  &   2078 & \textbf{78116} &\textbf{0}\\
\hline
Normal frames &   0  & 25804 &  19722 &  18041 &  60420 & 31525 &  7855 &  10922 &  4512 &  65165 &  68609 &   13804 &   8999 & \textbf{335378} & \textbf{485227}\\

\end{tabular}}
\caption{The first three rows display the number of videos, HR-videos and HR-test videos per category in UCF-Crime dataset, respectively. The number of anomalous and normal frames in HR-Crime for testing is shown in the bottom two rows. None of the \textit{Abuse} videos is annotated with ground truth labels by the authors of \cite{UCF_Crime}.
} %}
\label{tab:number_of_crime_videos}
\end{table}

The resulted HR-Crime dataset consists of 789 human-related anomaly videos and 782 human-related normal videos. Examples are shown in Figure \ref{fig:crime_dataset_examples}.
%We also annotated each video with a short description explaining the content of it, similar to the sub-captions of Figure \ref{fig:crime_dataset_examples}.
Table \ref{tab:number_of_crime_videos} shortly describes UCF-Crime and the newly proposed HR-Crime datasets. HR-Crime consists of 239 testing videos with annotation. Each video frame in this test set is annotated as normal or anomalous. As it can be observed, most categories consist mainly out of human-related videos. For instance, for \textit{Shoplifting}, all the videos are human-related. In contrast, \textit{Road accidents} has relatively the least number of human-related videos, which is expected as the people are mostly in cars, hence, not visible in the cameras.

Figure \ref{fig:video_length} compares the range and distribution of the video length in minutes for different categories. We observed that the video length varies for all categories. \textit{RoadAccidents} has the smallest variability in length compared to the other categories. In contrast, videos from the categories \textit{Shoplifting}, \textit{Arrest}, \textit{Fighting} and \textit{Burglary} vary the most in length. However, the \textit{Normal} category has the most number of videos that are longer than other videos of the same category. Inspecting the HR-Crime dataset, we realized that the longest \textit{Normal} video lasts 93.62 minutes.

A comparison of the range and distribution of the number of tracked skeletons is shown in Figure \ref{fig:number_of_people_detected}.
\textit{Arson} varies the least in number of detected people. In contrast, \textit{Arrest}, \textit{Normal}, \textit{Shoplifting} and \textit{Fighting} range the most. After further analysis we realized that the \textit{Normal} category has a video with a maximum of 1084 skeleton trajectories.

%%%%%%%% --------------------------------------
%%%%%%%% --------------------------------------

\subsection{Feature extraction}
The feature extraction pipeline starts with extracting human proposals from each frame of the video and extract skeletons from each proposal. As shown in Figure \ref{fig:pipeline}, first, human body proposals are detected using YOLOv3-spp, a Dense Connection and Spatial Pyramid Pooling Based version of YOLOv3 \cite{YOLOv3-spp}. Out of human body proposals, skeletons are retrieved employing AlphaPose \cite{alphapose}. Each skeleton consists of 17 keypoints representing body part locations. Skeletons are later tracked using PoseFlow \cite{poseflow} which uses FLANN and ORB descriptors to match two consecutive frames. The model we employed then uses tracked information in the last default 100 frames and using Hungarian algorithm \cite{Hungarian_method} solves the bipartite matching problem for the current frame. PoseFlow uses Non-Maximum Suppression just as AlphaPose to remove redundant trajectories and rematch disjoint pose flows. Given an input video, the skeletons are tracked over the frames to model the trajectory of each person appearing in a given video. The trajectories are later converted to CSV-files, which can be inputted to MPED-RNN human-based anomaly detector.

%%%%%%%% --------------------------------------
%%%%%%%% ------------ Experiments  --------
%%%%%%%% --------------------------------------
%%%%%%%% --------------------------------------

\section{Experiments}
\label{Section:results}
To evaluate MPED-RNN, authors in \cite{skeleton_based_anomaly_detection} mainly used the HR-ShanghaiTech dataset \cite{skeleton_based_anomaly_detection}. This dataset originally does not contain ground-truth labels for human bounding boxes, skeletons or trajectories. Also, the feature extraction step on HR-ShanghaiTech is missing from the MPED-RNN pipeline provided by the authors. Therefore, we were unable to evaluate the feature extraction steps separately on the HR-ShanghaiTech. Thus, we propose to compare the performance of the pre-trained MPED-RNN on the trajectories obtained using our feature extraction pipeline against the trajectories provided by the authors. For completeness of the proposed dataset, we also include the baseline results of the MPED-RNN network applied on the skeleton trajectories of the HR-Crime.

\subsection{Datasets}
\label{Subsection:data sets}

The HR-ShanghaiTech \cite{skeleton_based_anomaly_detection} consists of 101 human-related videos with anomalies such as running, jumping, riding a bicycle, etc. against the normality which is simply walking. HR-ShanghaiTech is a subset of ShanghaiTech dataset \cite{ShanghaiTech_data} consisting of 107 videos filmed by 12 cameras on the Shanghai Tech University campus.% with various types of anomalies.

Both the HR-ShanghaiTech test set \cite{skeleton_based_anomaly_detection} and UCF-Crime \cite{UCF_Crime} test set are provided with frame-level ground truth labels by the authors indicating if the event is anomalous or normal. Table \ref{tab:number_of_crime_videos} shows that 239 videos of the UCF-Crime testing set are human-related (the sum of 117 anomalous and 122 normal videos). Only these videos are used for the evaluation since the remaining videos are not labelled at frame-level.

%%%%%%%% --------------------------------------
%%%%%%%% ----------------- Results ------------
%%%%%%%% --------------------------------------
%%%%%%%% --------------------------------------

\subsection{Results and discussion}

Following \cite{skeleton_based_anomaly_detection}, we use the frame-level Receiver Operating Characteristic (ROC) curve and its corresponding Area Under (AUROC) to evaluate the performance of the methods on the HR-ShanghaiTech and HR-Crime datasets. A higher AUROC value indicates better performance. 

\begin{figure}[H]
    \centering
    \begin{subfigure}[b]{\textwidth}
         \centering
         \begin{subfigure}[b]{0.25\textwidth}
             \centering
             \includegraphics[width=\textwidth]{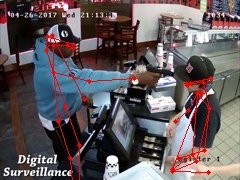}
             %\caption{Burglary}
         \end{subfigure}
         \hfill
         \begin{subfigure}[b]{0.25\textwidth}
             \centering
             \includegraphics[width=\textwidth]{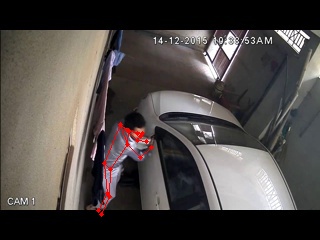}
             %\caption{Stealing}
         \end{subfigure}
         \hfill
         \begin{subfigure}[b]{0.25\textwidth}
             \centering
             \includegraphics[width=\textwidth]{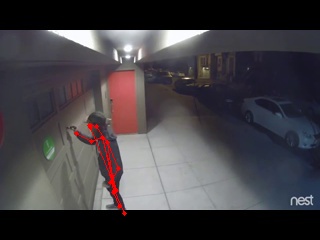}
             %\caption{Vandalism}
         \end{subfigure}
         %\label{fig:HR-Crime results}
         \caption{Satisfactory feature extraction}
    \end{subfigure}
    \newline
    \begin{subfigure}[b]{\textwidth}
     \centering
     \begin{subfigure}[b]{0.25\textwidth}
         \centering
         \includegraphics[width=\textwidth]{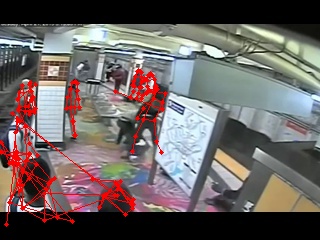}
     \end{subfigure}
     \hfill
     \begin{subfigure}[b]{0.25\textwidth}
         \centering
         \includegraphics[width=\textwidth]{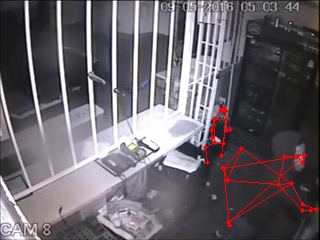}
     \end{subfigure}
     \hfill
     \begin{subfigure}[b]{0.25\textwidth}
         \centering
         \includegraphics[width=\textwidth]{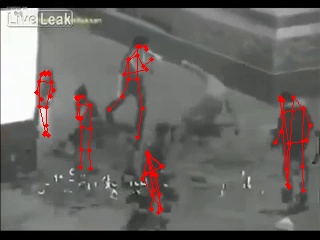}
     \end{subfigure}
     \caption{Unsatisfactory feature extraction}
     \label{fig:HR-Crime results issues}
\end{subfigure}
\vspace{-.2in}
    \caption{Visual examples of skeletons (in red) obtained from HR-Crime videos using our feature extraction pipeline. The first row displays examples of satisfactory extractions from clear videos. The bottom row shows examples of unsatisfactory extractions.} %\maya{Kayleigh, please change the sub-captions of this figure to only (a) and (b). (a) for first row, satisfactory feature extraction, (b) unsatisfactory. Also, for the second row, do you have examples of some other videos?}}
    \label{fig:HR-Crime results}
\end{figure}

\textit{A) Pose extraction HR-Crime:} Figure \ref{fig:HR-Crime results} shows a few examples of skeletons extracted from HR-Crime using our feature extraction pipeline discussed in Section \ref{Section:methodology}. As can be seen, the skeletons are reasonably accurate for higher quality videos and where the person is clearly in the camera view. However, for videos of lower quality or insufficient lighting, such as in Figure \ref{fig:HR-Crime results issues}, the feature extraction pipeline fails to detect people and their poses.

\textit{B) New baseline for HR-ShanghaiTech:} Authors of  \cite{skeleton_based_anomaly_detection} made available 12 models, each trained separately on a camera subdivision of HR-ShanghaiTech. HR-Crime however, is not structured in a camera wise manner. Therefore to ensure further consistencies, we trained the MPED-RNN architecture, \textit{de novo}, on the whole HR-ShanghaiTech training set from all the 12 cameras. For this training, we still used the trajectories provided by the authors. The model obtained from training on the entire HR-ShanghaiTech training set achieves a slightly lower performance in AUROC: 0.735 (see Table \ref{table-results1}) compared to the 0.754 reported in \cite{skeleton_based_anomaly_detection}, where 12 models are trained on the 12 camera subdivisions of HR-ShanghaiTech. This indicates that camera settings influence the complexity of the anomaly detection problem, at least when using MPED-RNN.

Moreover, as the authors of MPED-RNN did not provide detailed information on feature extraction steps from the videos, it is not possible to determine if feature extraction was purely done automatically or if human input was involved in refining the extracted features. This might be an important reason why using our feature extraction pipeline performs less accurately than the original trajectories given in \cite{skeleton_based_anomaly_detection} (0.534 AUROC score as compared to 0.735 AUROC score obtained by \cite{skeleton_based_anomaly_detection}).

\textit{C) MPED-RNN on HR-Crime:} Our first approach to establish the HR-Crime baseline is to test the pre-trained MPED-RNN with the entire HR-ShanghaiTech training set (on trajectories provided by \cite{skeleton_based_anomaly_detection}), on the HR-Crime test set, without explicitly fine-tuning the model to the new domain. We report the performance on HR-Crime class-wise as well as on the entire test set. This shows how well the information learned from the HR-ShanghaiTech can generalize to other categories of crimes scenes.

The obtained results are presented in Table \ref{table-results1} and Table \ref{table-results2}. They are reported based on the frame level reconstruction and prediction anomaly scores obtained by the MPED-RNN models. 
%We report the frame level performance measured on trajectory reconstructions and predictions based on anomaly scores. 
We observe that the AUROC scores are highest for videos of the type \textit{Assault} (0.7487) and \textit{Stealing} (0.7337). In contrast, \textit{Arson} and \textit{RoadAccidents} have the lowest AUROC scores, 0.4290 and 0.4171 respectively. These results indicate that the pre-trained MPED-RNN model can make promising predictions even on an unseen domain if the `human' subjects are fairly present in committing the anomalies. This is also an indication of suitability of the proposed feature extraction pipeline.

%% ---------%% ---------%% ---------%% ---------%% ---------%% ---------%% ---------
%% ---------%% ---------%% ---------%% ---------%% ---------%% ---------%% ---------
%% ---------%% ---------%% ---------%% ---------%% ---------%% ---------%% ---------

\begin{table}[!htb]
\begin{minipage}{.5\linewidth}
\centering
\begin{tabular}{ll|c}
\multicolumn{2}{l|}{Dataset} & AUROC \\ 
\hline \hline
\multicolumn{2}{l}{\textit{B) New baseline for HR-ShanghaiTech}}\\
\hline
\multicolumn{2}{l|}{HR-ShanghaiTech} & \textbf{0.7346}  \\
\hline
\multicolumn{2}{l}{\textit{C) MPED-RNN on HR-Crime}}\\
\hline
& HR-ShanghaiTech (pre-trained) \cite{skeleton_based_anomaly_detection} & 0.5346    \\
\hline
\multicolumn{2}{l}{\textit{D) Fine-tuning MPED-RNN on HR-Crime}}\\
\hline
\multicolumn{2}{l|}{HR-Crime (de novo)} & \textbf{0.6030}  \\
\multicolumn{2}{l|}{HR-ShanghaiTech (fine-tuned)} & 0.5879  \\
\end{tabular}
\vspace{0.1in}
\caption{Baseline results for the HR-ShanghaiTech and HR-Crime datasets using MPED-RNN models.}
\label{table-results1}
\end{minipage}%
\begin{minipage}{.5\linewidth}
\centering
\begin{tabular}{c|c|c}
Dataset & Crime class & AUROC \\
\hline \hline
\multicolumn{3}{l}{\textit{C) MPED-RNN on HR-Crime}}\\
\hline

      \multirow{13}*{\rotatebox[origin=l]{90}{HR-ShanghaiTech (pre-trained)\cite{skeleton_based_anomaly_detection}}}& Arrest         & 0.5617   \\
& Arson          & 0.4290    \\
& Assault        & 0.7487     \\
& Burglary       & 0.6790   \\
& Explosion      & 0.4740     \\ 
& Fighting       & 0.4847    \\ 
& Road accidents & 0.4171     \\
& Robbery        & 0.6586    \\
& Shooting       & 0.4900    \\
& Shoplifting    & 0.6342   \\
& Stealing       & 0.7337    \\
& Vandalism      & 0.6396    \\
\end{tabular}
\vspace{0.1in}
\caption{Class-wise results for the HR-Crime dataset using the pre-trained MPED-RNN on HR-ShanghaiTech.}
\label{table-results2}
\end{minipage} 
\end{table}
%%%%%%%% --------------------------------------
%%%%%%%% --------------------------------------
%%%%%%%% --------------------------------------
%%%%%%%% --------------------------------------

%\noindent\textit{Approach 2: adapt model to HR-Crime:} 
\textit{D) Fine-tuning MPED-RNN on HR-Crime:} 
The second approach for establishing the HR-Crime baseline is training the MPED-RNN architecture to the newly created HR-Crime. For this purpose, we train the MPED-RNN model \textit{de novo} on the HR-Crime dataset. For completeness, we also propose to fine-tune the pre-trained MPED-RNN model on the HR-ShanghaiTech dataset, on HR-Crime. With this approach, we aim at closing the structural gap between the two datasets. The results are presented in Table \ref{table-results1}. The model trained \textit{de novo} on HR-Crime achieves 0.6030 AUROC performance which is explained by the increased complexity of the HR-Crime dataset as compared to HR-ShanghaiTech. Surprisingly, the fine-tuned model achieves a lower AUROC performance of 0.5879 as compared to the model trained \textit{de novo}. We suspect, the information gain from the HR-ShanghaiTech model does not generalize well to the HR-Crime dataset. This emphasizes the complexity gap between the two datasets that is apparent when comparing anomalies such as running and jumping in HR-ShanghaiTech to the anomalous events in HR-Crime. Comparing to the results of the experiment (C), our fine-tuned model shows a slight increase in performance which might also suggest that the pre-training approach is relevant to consider when comparing two datasets with distinct structures.

\section{Conclusion}\label{Section:discussion}

In this work, we discussed the preparation steps to develop HR-Crime, and provided a baseline human-related anomaly detection analysis on it. The dataset as well as the feature extraction pipeline will be publicly available for further use by the community.

The results presented in Section \ref{Section:results} (C) show that the pre-trained MPED-RNN on HR-ShanghaiTech does not perform as well on HR-Crime compared to the performance achieved on HR-ShanghaitTech itself. We suspect this to happen due to the complexity gap between the two datasets as also has been observed in other domains \cite{indoor_scenes} and \cite{places}. HR-ShanghaiTech consists of videos shot on the same University campus. HR-Crime, on the other hand, is a collection of YouTube videos, where each video is filmed in a different location. Also, the types of anomalies differ greatly: HR-ShanghaiTech contains anomalous events such as \textit{running} and \textit{jumping}, while HR-Crime consists of real-world crime scenes with natural movements that are not staged. The quality of the videos also plays an important role in anomaly detection. The HR-ShanghaiTech videos are of high quality, while the HR-Crime videos range in quality and lighting. Thus, skeletons and trajectories are detected less accurately for HR-Crime.

Another factor that plays an essential role in the high number of false negatives for anomalous events is the frame-level evaluation. Each test video in HR-Crime is annotated with ground truth labels indicating the window of an anomalous event. These windows indicate the start and end frame of the event. As mentioned before, the HR-Crime dataset is composed of a complex set of videos. Therefore an anomalous event such as a \textit{Burglary} can have multiple anomalous and normal movements in the same video. However, prediction and evaluation are made per frame. An alternative that we plan on exploring is to label the videos temporally and spatially to not only evaluate if a frame contains an anomalous event but also to find the area where it occurs as suggested in \cite{streetscene}. This is a more accurate way of evaluating anomalous events. However, it requires laborious work to annotate video frames manually. Future lines of research will also be dedicated to the categorical classification of the identified anomalies through the analysis of the descriptors of the movement of the human body.

\bibliographystyle{splncs04}
\bibliography{references}

\begin{thebibliography}{10}
\providecommand{\url}[1]{\texttt{#1}}
\providecommand{\urlprefix}{URL }
\providecommand{\doi}[1]{https://doi.org/#1}

\bibitem{openpose}
Cao, Z., Hidalgo, G., Simon, T., Wei, S.E., Sheikh, Y.: Openpose: realtime
  multi-person 2d pose estimation using part affinity fields. IEEE Transactions
  on PAMI  (2019)

\bibitem{human_anomaly}
Emonet, R., Varadarajan, J., Odobez, J.M.: Multi-camera open space human
  activity discovery for anomaly detection. IEEE International Conference on
  AVSS  (2011)

\bibitem{alphapose}
Fang, H.S., Xie, S., Tai, Y.W., Lu, C.: Rmpe: Regional multi-person pose
  estimation. IEEE International Conference on Computer Vision  (2017)

\bibitem{human_anomaly_1}
Gong, M., Zeng, H., Xie, Y., Li, H., Tang, Z.: Local distinguishability
  aggrandizing network for human anomaly detection. Neural Networks
  \textbf{122},  364--373 (2020)

\bibitem{YOLOv3-spp}
Huang, Z., Wang, J., Fu, X., Yu, T., Guo, Y., Wang, R.: Dc-spp-yolo: Dense
  connection and spatial pyramid pooling based yolo for object detection.
  Information Sciences  (2020)

\bibitem{arttrack}
Insafutdinov, E., Andriluka, M., Pishchulin, L., Tang, S., Levinkov, E.,
  Andres, B., Schiele, B.: Arttrack: Articulated multi-person tracking in the
  wild. IEEE Conference on Computer Vision and Pattern Recognition  (2017)

\bibitem{deepercut}
Insafutdinov, E., Pishchulin, L., Andres, B., Andriluka, M., Schiele, B.:
  Deepercut: A deeper, stronger, and faster multi-person pose estimation model.
  European Conference on Computer Vision  (2016)

\bibitem{posetrack}
Iqbal, U., Milan, A., Gall, J.: Posetrack: Joint multi-person pose estimation
  and tracking. IEEE Conference on Computer Vision and Pattern Recognition
  (2017)

\bibitem{metu}
Kocabas, M., Karagoz, S., Akbas, E.: Multiposenet: Fast multi-person pose
  estimation using pose residual network. European conference on computer
  vision  (2018)

\bibitem{Hungarian_method}
Kuhn, H.W.: The hungarian method for the assignment problem. Naval Research
  Logistics Quarterly pp. 83--97 (1955)

\bibitem{ShanghaiTech_data}
Liu, W., Luo, W., Lian, D., Gao, S.: Future frame prediction for anomaly
  detection--a new baseline. IEEE Conference on Computer Vision and Pattern
  Recognition  (2018)

\bibitem{skeleton_based_anomaly_detection}
Morais, R., Le, V., Tran, T., Saha, B., Mansour, M., Venkatesh, S.: Learning
  regularity in skeleton trajectories for anomaly detection in videos. IEEE
  Conference on Computer Vision and Pattern Recognition  (2019)

\bibitem{indoor_scenes}
Quattoni, A., Torralba, A.: Recognizing indoor scenes. In: 2009 IEEE Conference
  on Computer Vision and Pattern Recognition. pp. 413--420. IEEE (2009)

\bibitem{streetscene}
Ramachandra, B., Jones, M.: Street scene: A new dataset and evaluation protocol
  for video anomaly detection. IEEE Winter Conference on Applications of
  Computer Vision  (2020)

\bibitem{video_anomaly_detection_survey}
Ramachandra, B., Jones, M., Vatsavai, R.R.: A survey of single-scene video
  anomaly detection. IEEE Transactions on Pattern Analysis and Machine
  Intelligence  (2020)

\bibitem{UCF_Crime}
Sultani, W., Chen, C., Shah, M.: Real-world anomaly detection in surveillance
  videos. IEEE Conference on Computer Vision and Pattern Recognition  (2018)

\bibitem{poseflow}
Xiu, Y., Li, J., Wang, H., Fang, Y., Lu, C.: Pose flow: Efficient online pose
  tracking. arXiv preprint arXiv:1802.00977  (2018)

\bibitem{places}
Zhou, B., Lapedriza, A., Khosla, A., Oliva, A., Torralba, A.: Places: A 10
  million image database for scene recognition. IEEE Transactions on Pattern
  Analysis and Machine Intelligence  \textbf{40}(6),  1452--1464 (2017)

\end{thebibliography}

\end{document}